\def\year{2020}\relax
\documentclass[letterpaper]{article} 
\usepackage{aaai21}  
\usepackage{times}  
\usepackage{helvet} 
\usepackage{courier}  
\usepackage[hyphens]{url}  
\usepackage{graphicx} 
\usepackage{graphicx}  
\usepackage{natbib} 
\usepackage{caption}
\usepackage{amssymb}
\usepackage{booktabs}
\usepackage{amsfonts}
\usepackage{algorithm}
\usepackage{algpseudocode}
\usepackage{amsmath}
\usepackage{amsthm}
\usepackage{graphics}
\usepackage{epsfig}
\usepackage{multirow}
\usepackage{subfigure}
\usepackage{bbm}
\usepackage{xcolor}
\usepackage{lipsum}

\usepackage{hyperref}
\newcommand{\printfnsymbol}[1]{%
  \textsuperscript{*}%
}

\title{MetaDelta: A Meta-Learning System for Few-shot Image Classification}

\author{Yudong Chen,\textsuperscript{*} \footnote{Equal contribution}
Chaoyu Guan,\textsuperscript{ *} 
Zhikun Wei,\textsuperscript{ *}
Xin Wang,\textsuperscript{$\dagger$} \footnote{Corresponding authors}
Wenwu Zhu\textsuperscript{$\dagger$}\\
{\rm  Department of Computer Science and Technology, Tsinghua University} \\
{\rm \textsuperscript{ *} \{cyd18, guancy19, weizk19\}@mails.tsinghua.edu.cn,}\\ 
{\rm \textsuperscript{$\dagger$} \{xin\_wang, wwzhu\}@tsinghua.edu.cn}
}
\begin{document}

\maketitle

\begin{abstract}
Meta-learning aims at learning quickly on novel tasks with limited data by transferring generic experience learned from previous tasks. Naturally, few-shot learning has been one of the most popular applications for meta-learning. 
However, existing meta-learning algorithms rarely consider the time and resource efficiency or the generalization capacity for unknown datasets, which limits their applicability in real-world scenarios. In this paper, we propose MetaDelta, a novel practical meta-learning system for the few-shot image classification. MetaDelta consists of two core components: i) multiple meta-learners supervised by a central controller to ensure efficiency, and ii) a meta-ensemble module in charge of integrated inference and better generalization. In particular, each meta-learner in MetaDelta is composed of a unique pretrained encoder fine-tuned by batch training and parameter-free decoder used for prediction. MetaDelta ranks first in the final phase in the AAAI 2021 MetaDL Challenge\footnote{ https://competitions.codalab.org/competitions/26638}, demonstrating the advantages of our proposed system. The codes are publicly available at https://github.com/Frozenmad/MetaDelta.
\end{abstract}

\section{Introduction}
\label{sec:introduction}

Despite the great success of machine learning, a clear gap between human and artificial intelligence is the ability to learn from small samples, e.g., learning to recognize objects from limited examples. Inspired by human's ability of learning to learn from experience, meta-learning~\cite{vanschoren2018meta} aims to transfer the generic experience learned from multiple tasks of limited data to efficiently complete new tasks. As one of the most successful applications for meta-learning, few-shot learning targets at learning from a limited number of labeled examples, which has become a research trend recently. Few-shot image classification is a task where the classifier must learn to accommodate new classes not seen during training with limited examples.

Existing meta-learning algorithms can be categorized into three groups: (1) \emph{Metric}-based methods~\cite{vinyals2016matching, snell2017prototypical} that learn an encoder for the examples and apply parameter-free inference (e.g., nearest neighbor) in the embedding space, (2) \emph{Optimization}-based methods~\cite{finn2017model, rusu2018meta, park2019meta} that extract the meta-knowledge of optimization algorithms for fast adaption, and (3) \emph{Black-box}- or \emph{Model}-based methods~\cite{santoro2016meta, mishra2017simple, munkhdalai2017meta} that directly learn to embed the datasets to model parameters for prediction. Among them, (1) and (2) have become the most popular methodologies and have been proved effective in various few-shot settings. However, there exist two challenges largely unexplored. First, most existing methods do not consider the time and resource efficiency or budget, which limits their ability to meet the requirement in many real-world applications. Furthermore, the success of existing methods heavily relies on careful hyperparameter designs (e.g., backbones, learning rates, etc.) on each specific dataset. In real-world scenarios, the datasets and tasks may be unknown, diverse, or even changing over time, making manually design of the most suitable hyperparameters very laborious. 

To tackle these challenges, we design a novel practical meta-learning system (MetaDelta) for few-shot image classification tasks in this paper. Following the metric-based methods, MetaDelta firstly adopts pretrained convolutional networks as backbones to project images to latent vectors and trains the backbones with linear classifiers in a non-episodic way on the training classes. To improve the system's generalization capacity to any unknown datasets under time and memory budgets, we employ multiple meta-learning models with multi-processing, while managing the time and resources with a central controller in the main process at the same time. Moreover, we implement a late-fusion meta-ensemble mechanism to improve the generalization ability by taking the prediction from each model into account. MetaDelta consistently outperforms competitive baselines on various datasets and ranks first in the final phase of AAAI 2021 MetaDL Challenge, which shows the superiority of our proposed meta-learning system.

\begin{figure*}[ht]
  \centering
  \includegraphics[width=0.6\linewidth]{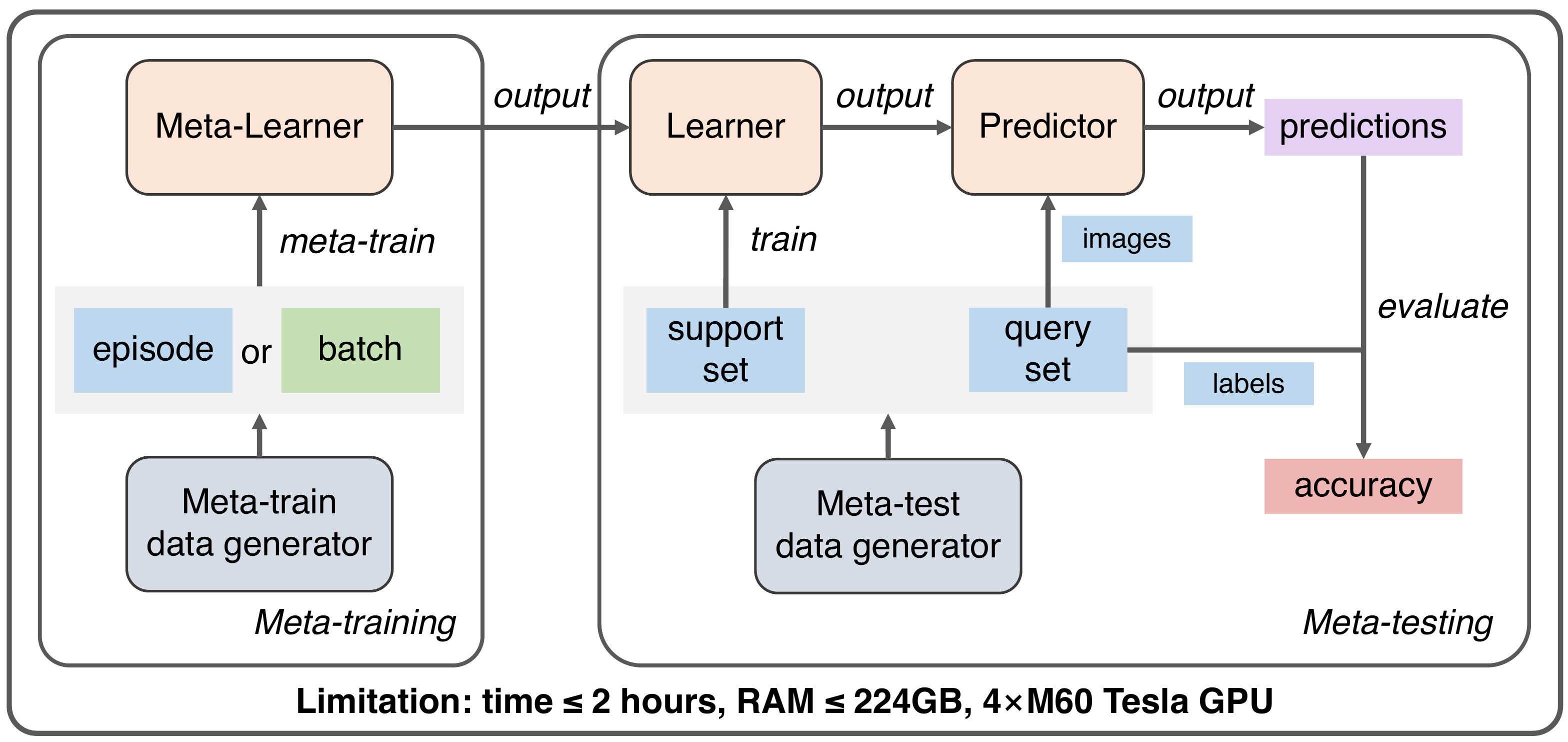}
  \caption{The workflow of the AAAI 2021 MetaDL Challenge on few-shot image classification.}
  \label{fig:workflow}
\end{figure*}

\section{AAAI 2021 MetaDL Challenge}
\label{sec:metadl_challenge}

In this section, we first introduce the workflow of a meta-learning system for the few-shot image classification task in AAAI 2021 MetaDL Challenge and then review the details and challenges of this competition.

As illustrated in Fig.~\ref{fig:workflow}, the workflow of such a meta-learning system is as follows. A meta-learner is first trained on the episodes or batches of data generated by the meta-train data generator. An \emph{episode} refers to a $K$-way $N$-shot image classification task $\mathcal{T} = \{\mathcal{D}_{spt}, \mathcal{D}_{qry}\}$, where the support set $\mathcal{D}_{spt}$ and the query set $\mathcal{D}_{qry}$ are the training set and the test set of this task, respectively. Note that $|\mathcal{D}_{spt}| = K N$, and $|\mathcal{D}_{qry}| = K Q$, where $K$ is the number of image categories of the task, and all $K, N, Q$ are manually adjustable before meta-training. A \emph{batch} refers to $B$ examples $\{\mathbf{x}_i, y_i\}$ randomly sampled from the meta-training data. The meta-learner outputs a learner, which is then trained on the support set of each meta-test episode to output a specific predictor for episodic evaluation. 

The AAAI 2021 MetaDL Challenge consists of a feedback phase and a final phase. During the feedback phase, an offline public dataset (the Omniglot dataset \cite{lake2015human}) and an online \emph{feedback dataset} (i.e., the dataset is unknown/unavailable and only used to evaluate submissions uploaded by participants) are provided for the participants to develop their meta-learning systems. During the final phase, new online feedback datasets are used to evaluate the submissions. The evaluation metric is the average classification accuracy on the query sets of 600 meta-test episodes. All the meta-test episodes are defined as 5-way 1-shot image classification tasks with an unknown number of query examples in each class (i.e., $K=5, N=1, Q$ is unknown).

As a challenge on meta-learning with few-shot image classification settings and online judge, this competition has the following challenges.

\begin{itemize}

\item \textbf{Fast adaption without overfitting.} This is the core challenge of meta-learning and is especially critical for few-shot learning settings. The 5-way 1-shot classification problem in the competition requires the meta-learners to learn generalized prior knowledge from meta-training tasks and a proper way for fast adaption on limited novel data without overfitting. 

\item \textbf{Time and resource efficiency.} Another challenging aspect of this competition is the efficiency requirements for the submissions: the whole meta-training and meta-testing workflow should be finished within 2 hours on an Azure NV24 machine with 4 M60 Tesla GPU and 224 GB of RAM. A superior meta-learning algorithm should thus not only learn fast (fast adaption), but also meta-learn fast. 

\item \textbf{Generalization across different datasets.} During both the feedback and final phases, the online feedback datasets are unknown/unavailable to participants. Therefore, a good submission must work well on any unknown dataset without manually tailored hyperparameters (e.g., learning rate, backbone structure, etc.). This is difficult since the image distribution of the feedback dataset may differ heavily from existing offline datasets with a different number of query examples, image size, etc. Furthermore, with the time and resource limitations, common AutoML methods such as hyperparameter optimization (HPO) and neural architecture search (NAS) can be too expensive to be applicable to automatically specify the best hyperparameters on the feedback/final dataset. 

\end{itemize}

\section{MetaDelta}
\label{sec:our_system}

In this section, we elaborate on our meta-learning system (MetaDelta) for the AAAI 2021 MetaDL Challenge.

\begin{figure*}[ht]
  \centering
  \includegraphics[width=0.7\linewidth]{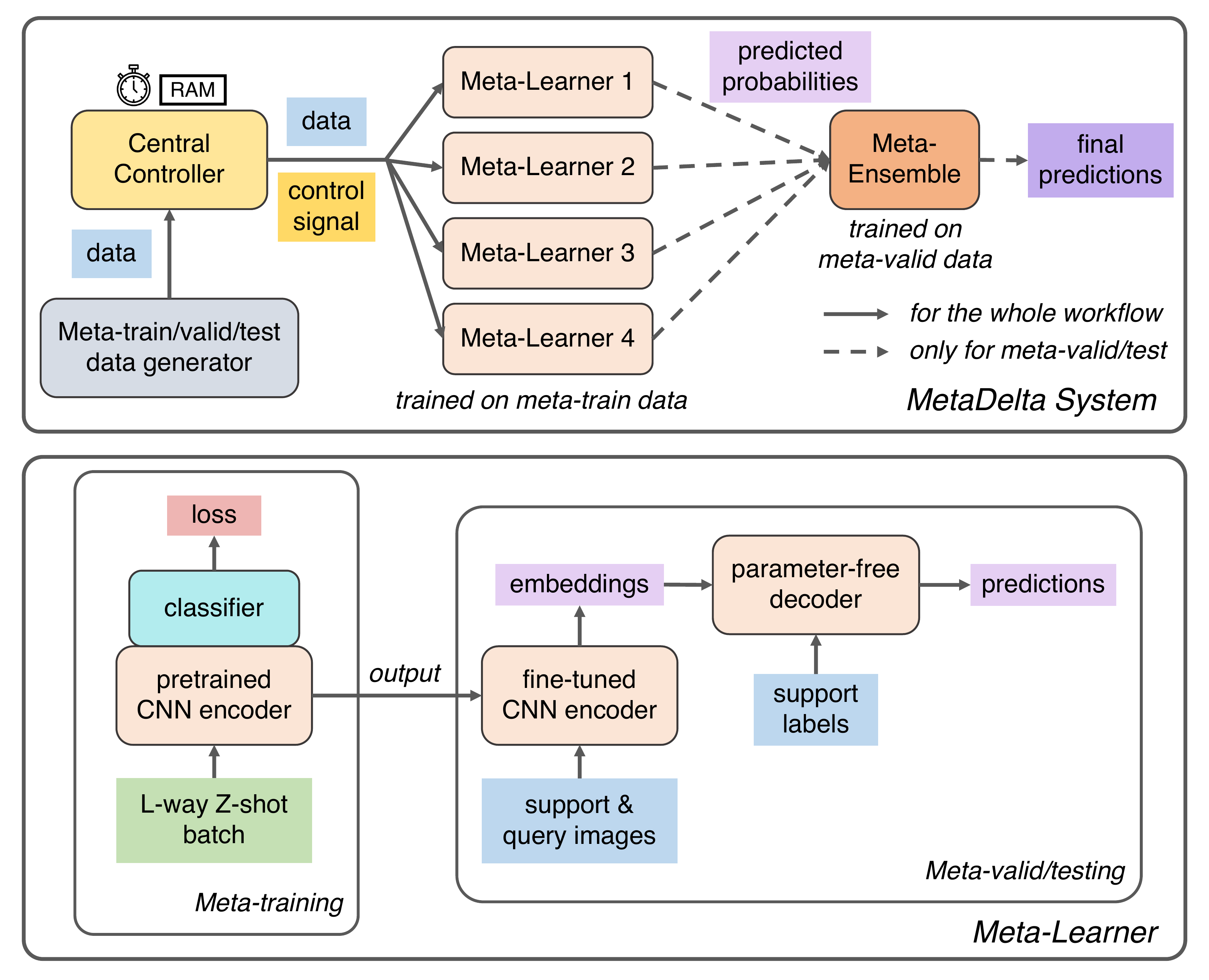}
  \caption{The MetaDelta system for the AAAI 2021 MetaDL Challenge. The top part depicts the whole multi-processing system with a central controller to ensure time and resource efficiency and a meta-ensemble module to improve generalization capacity to unknown datasets. The bottom part shows the framework of the Meta-Learners in MetaDelta, such that different Meta-Learners can be instantiated with different compositions of their components. CNN = Convolutional Neural Network.}
  \label{fig:our_system}
\end{figure*}

\subsection{System Overview}
\label{subsec:system_overview}

The MetaDelta system is illustrated in Fig~\ref{fig:our_system}. To tackle the challenge of time and resource efficiency, we adopt a central controller in the main process to dispatch data and decide when to start and stop the meta-training/testing (top of Fig~\ref{fig:our_system}). Aiming to achieve good and robust performances on unknown feedback datasets, a meta-ensemble is learned to ensemble 4 different meta-learners. The 4 meta-learners are derived by training with different hyperparameters parallelly on 4 GPUs, which is managed by the central controller.

A specific meta-learner in MetaDelta is instantiated and used following the framework illustrated in the bottom of Fig~\ref{fig:our_system}, which is capable of fast adaption in the few-shot image classification problem. During the meta-training period, we leverage batch training strategy to train a deep model to classify all the meta-training classes (e.g., if the meta-training set includes 500 classes, then the model applies 500-way classification). For the sake of time efficiency and generalization capacity to unknown datasets, we leverage universal pretrained CNN encoders (e.g., ResNet50 pretrained on ImageNet) to embed images into features, and add a classifier head onto the encoder for fine-tuning. During the meta-testing period, we discard the classifier head and map the images to embeddings with the fine-tuned encoder, and apply an efficient parameter-free decoder to predict the class labels of query images based on the embeddings. The optimal meta-learner components are selected based on experimental evaluations on various offline datasets.

\subsection{Meta-Learners}

As aforementioned in Sec~\ref{sec:introduction}, \emph{Metric} based (e.g., Prototypical Networks, a.k.a. ProtoNet~\cite{snell2017prototypical}) and \emph{Optimization} based (e.g., MAML~\cite{finn2017model}) methods are the most popular and effective meta-learners in existing literature. Through extensive experiments on the 5-way 1-shot image classification tasks, we find ProtoNet-like methods outperform MAML-like methods on various datasets, and thus select ProtoNet-like frameworks as our meta-learners (the bottom of Fig~\ref{fig:our_system}). 

Different from ProtoNet, we apply non-episodic (i.e., batch) training instead of episodic training to learn a CNN encoder to embed images to feature vectors, since we find non-episodic training leads to more effective encoder in our experiment (See. Table \ref{tab:main_results}). Then a parameter-free decoder is taken during meta-valid/testing periods to decode the vectors of each episode to the predicted labels. 
In detail, we iteratively train and evaluate the CNN encoder: First, the encoder will be meta-trained $r$ epochs using non-episodic training strategy. Then, the episodic classification accuracy will be calculated using parameter-free decoder on meta-valid dataset. The model with the best meta-valid episodic accuracy will be saved for further use.

\subsubsection{Fine-tuned CNN Encoder}

We select pretrained CNN backbones (pretrained on ImageNet \cite{deng2009imagenet}) as the initialized encoder and then fine-tune it on batches of meta-training data. We use pretrained deep backbones since they have strong generalization capacity and help the meta-learner to generalize to unknown feedback datasets. To some degree, the pretrained models can also be regarded as a meta knowledge collection of ImageNet. Moreover, compared to learning from scratch, fine-tuning on pretrained CNNs also saves time and computing resources, enabling effective training of powerful deep models within the time limit. In our experiment, we select ResNet50 \cite{he2016deep}, ResNet152 \cite{he2016deep}, WRN50 \cite{zagoruyko2016wide} and MobileNet \cite{sandler2018mobilenetv2} for the four meta-learners in MetaDelta. 

To fine-tune the backbones, a linear classifier head (i.e., fully-connected layer) is added to the final layer of the CNN encoder, and we randomly sample $L$-way $Z$-shot batches from the meta-train classes for training. Here, $L$-way $Z$-shot means each batch consists of $L$ classes with $Z$ labeled examples for each class, which is set to keep class balance. To augment the image data and learn a more robust encoder, we follow \cite{chen2019self} to apply the rotation loss: First, we rotate each image in a batch by $0, 90, 180, 270$ degrees to get four images, which should be classified into the same class by the original classifier head. Then, another 4-way linear classifier head is added to the top of the CNN encoder to predict the four kinds of rotations. Finally, we optimize the weights of encoders by minimizing the following loss:
$$L=L_{cls} + \alpha L_{rot}$$
where $L_{cls}$ is the classification loss, $L_{rot}$ is the rotation loss, and $\alpha$ is a hyperparameter for balance.

\subsubsection{Parameter-free Decoder}

With the feature vectors encoded by the fine-tuned CNN encoder, we could further predict the labels of query examples in meta-valid/testing episodes with the help of parameter-free decoders. During meta-valid period, we use the decoder in ProtoNet~\cite{snell2017prototypical} to make inference. The models with the best few-shot classification accuracy on meta-valid dataset is chosen as the encoder for further use.

Specifically, given a meta-valid episode of $N$-way $K$-shot, we first compute the prototypes from the support set $\mathcal{S}_j$ of the $j$-th class:
\begin{equation}
    \small
	\label{eqn:prototypes}
	\mathbf{c}_j = \frac{1}{K} \sum_{(\mathbf{x}_i, y_i) \in \mathcal{S}_{j}} f_{\phi} (\mathbf{x}_i),
\end{equation}
\noindent where $f_{\phi}$ denotes the CNN encoder. Then, the ProtoNet decoder produces a distribution over classes for each query example $\mathbf{x}$ based on a softmax over the Euclidean distances between its embedding and the $K$ prototypes:
\begin{equation}
    \small
	\label{eqn:prototype_softmax}
	p_{\phi}(y = j | \mathbf{x}) = \frac{\exp (- d(f_{\phi}(\mathbf{x}), \mathbf{c}_j))}{\sum_{j'} \exp (- d(f_{\phi}(\mathbf{x}), \mathbf{c}_{j'}))},
\end{equation}
\noindent where $(\mathbf{x}, y)$ is a query example, and $d(\mathbf{a}, \mathbf{b})$ denotes Euclidean distance between vectors $\mathbf{a}$ and $\mathbf{b}$. The prediction is then made by classifying the example to the most probable class. 

During meta-test period, we implement the soft $k$-means based transductive decoder in MCT~\cite{kye2020transductive} to build more accurate prototypes by considering query embeddings. Concretely, the initial prototypes $\{\mathbf{c}_j^{(0)}\}$ are the same as that in Eq.~\ref{eqn:prototypes}. The MCT decoder iteratively updates the prototypes for $T$ steps. For each step $t$, we first calculate the confidence scores $q_j^{(t-1)}(\mathbf{x})$ for each query example $\mathbf{x}$ belonging to class $j$ in the same way as Eqn.~\ref{eqn:prototype_softmax}:
\begin{equation}
    \small
	\label{eqn:mct_confidence_score}
	q_j^{(t-1)}(\mathbf{x}) = \frac{\exp (- d(f_{\phi}(\mathbf{x}), \mathbf{c}_j^{(t-1)}))}{\sum_{j'} \exp (- d(f_{\phi}(\mathbf{x}), \mathbf{c}_{j'}^{(t-1)}))}.
\end{equation}
\noindent Then, we update the prototypes based on the confidence scores for all query examples $\mathbf{x}$ in episodic query set $\mathcal{Q}$:
\begin{equation}
    \small
	\label{eqn:mct_prototype_update}
	\mathbf{c}_j^{(t)} = \frac{ \sum_{\mathbf{x}\in \mathcal{S}_j} 1 \cdot f_{\phi}(\mathbf{x}) + \sum_{\mathbf{x}\in \mathcal{Q}} q_j^{(t-1)}(\mathbf{x}) \cdot f_{\phi}(\mathbf{x}) }{ \sum_{\mathbf{x}\in \mathcal{S}_j} 1 + \sum_{\mathbf{x}\in \mathcal{Q}} q_j^{(t-1)}(\mathbf{x}) }.
\end{equation}
\noindent The predictions are finally made based on $q_j^{(T)}(\mathbf{x})$.\footnote{We do not use the learnable distance metric proposed in MCT, which brings no improvement in our preliminary experiments.}

We find in our experiments that the accuracy trendlines of the ProtoNet decoder and MCT decoder are generally the same (see Fig~\ref{fig:decoder_cmp}), while the latter leads to higher accuracies given the same CNN encoder. Therefore, we take the ProtoNet decoder for meta-valid to accelerate training without missing the best models, and the MCT decoder for meta-test to make more accurate inferences. 
Note that using a decoder during meta-validation to calculate episodic accuracies (instead of batch-wise classification accuracies as during meta-training) is reasonable, since a CNN encoder that facilitates low meta-training loss does not ensure high episodic accuracy during meta-valid/test periods.


\subsection{Meta-Ensemble}

The meta-ensemble module is designed to tackle the challenge of generalization capacity, i.e., to improve the performance of MetaDelta on any unknown feedback/final dataset. Ensemble methods have been empirically proved to be effective in various supervised classification tasks~\cite{rokach2010ensemble}. In MetaDelta, the meta-ensemble module integrates the predicted probabilities of the four meta-learners and outputs the final predictions, as illustrated in Fig.~\ref{fig:our_system}. 

The meta-ensemble model is trained after finishing the meta-training of all meta-learners. To train the meta-ensemble model, we divide the meta-valid data into a training set $\mathcal{D}_{val}^{tr}$ and a test set $\mathcal{D}_{val}^{te}$. Taking the concatenation of the predicted probabilities from the four loaded best meta-learners as input, several meta-ensemble models are trained on $\mathcal{D}_{val}^{tr}$ simultaneously and evaluated on $\mathcal{D}_{val}^{te}$ based on episodic accuracy. The best meta-ensemble model is then saved for the inference in the meta-test period. In our experiments, we implement voting, Gradient Boost Machine, General Linear Model, Naive Bayesian Classifier, and Random Forest\footnote{We leverage the Gradient Boost Machine library from a lib \textbf{LightGBM} (\url{lightgbm.readthedocs.io/en/latest/index.html}). Naive Bayesian Classifier and Random Forest models are implemented based on sklearn \cite{scikit-learn}.} as the meta-ensemble candidate models. Due to the diversity of suitable scenarios of these models, we argue that our meta-ensemble module is capable of dynamically adapting to the unknown feedback dataset by selecting the best ensemble model according to the meta-valid data. This design further improves the robustness of our system.

\subsection{Central Controller}

The central controller module aims at improving the time and resource efficiency of our system and avoiding timeout or memory overrun. All the multi-thread and multi-processing designs in this module facilitate our system to support a greater degree of parallelism and make full use of the computing resources. 

\begin{figure}[ht]
  \centering
  \includegraphics[width=1\linewidth]{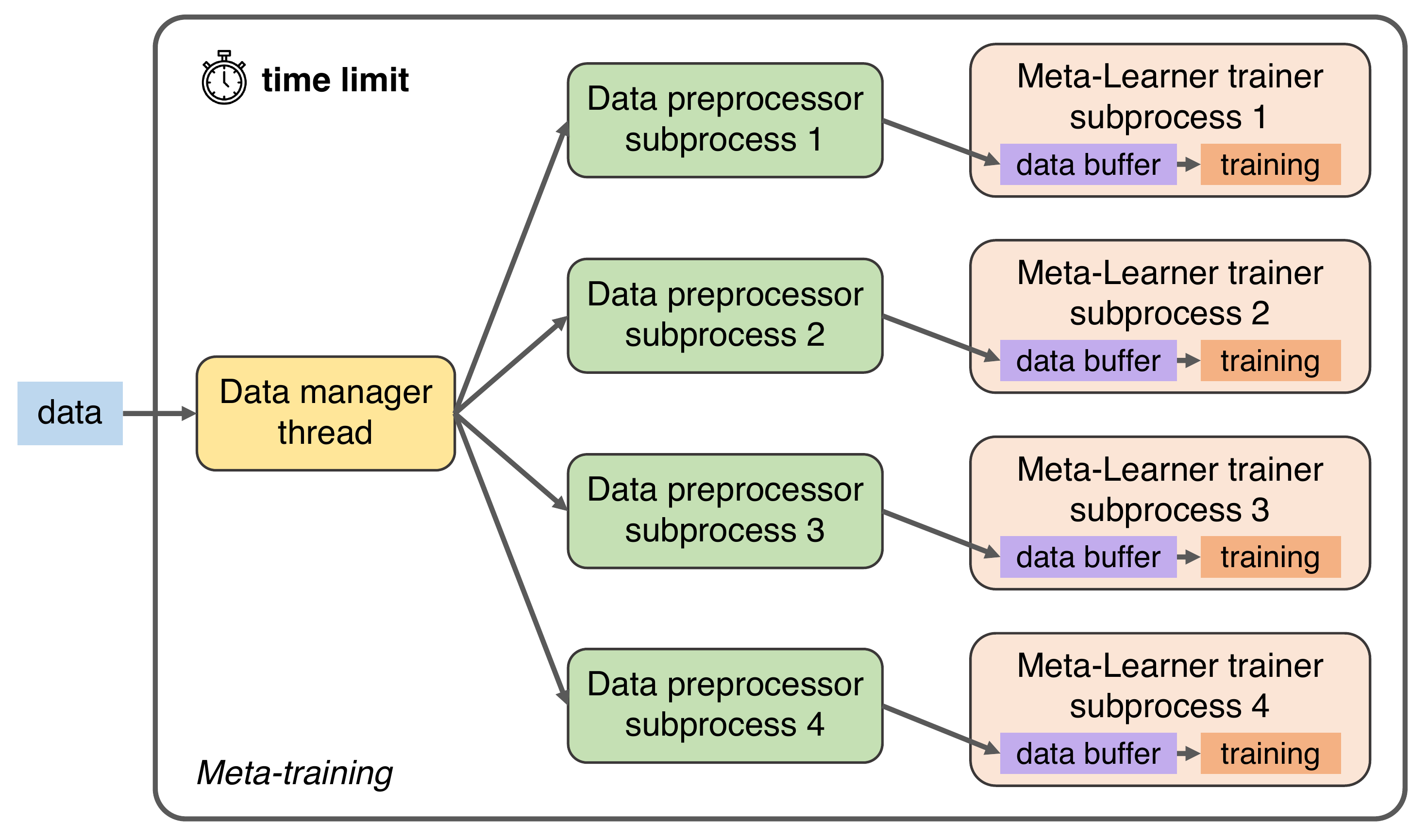}
  \caption{The central controller in MetaDelta that leverages multi-thread and multi-processing techniques to improve time and resource efficiency and support meta-ensemble.}
  \label{fig:central_controller}
  \vspace{-1mm}
\end{figure}

The design of central controller is illustrated in Fig.~\ref{fig:central_controller} (during meta-training period). First, a timer is set in the main process to measure and estimate the time cost of meta-training/testing epochs, based on which the whole meta-training/testing procedure is supervised. A subprocess may be  killed in advance by the central controller if it is predicted to run out of time budget. Under this framework, the main process starts a data manager thread to load, copy, and dispatch the episode/batch data. Then, four data preprocessor subprocesses are started to receive and preprocess the data copies according to the requirements of specific meta-learner trainer subprocesses. The preprocessed data is sent to the corresponding data buffer, which is designed to support asynchronous meta-training in different subprocesses. The data preprocessor subprocess will not sleep until timeout or all the buffers are full. Each meta-learner trainer subprocess is run on one GPU, and the main process and data preprocessor subprocesses are run on CPU and the RAM. 

During meta-valid and meta-test periods, the same central controller framework is used except for the training and inference of the meta-ensemble module in the main process.

\section{Experiments}

In this section, we demonstrate our ranking in the final phase and experimental results in offline evaluations. 

\subsection{Final Phase Results}

AAAI 2021 MetaDL Challenge consists of a feedback phase and a final phase, which phases take different unknown online feedback datasets to test the submissions of participants. The ranks of the top 3 teams in the final phase are shown in Table~\ref{tab:ranking}. Our team ranks first with a meta-test accuracy of 0.4042, which validates the effectiveness and generalization capacity of MetaDelta.

\begin{table}[ht]
  \centering
  \caption{Rankings of the top 3 teams in the final phase of the AAAI 2021 MetaDL Challenge.}
  \label{tab:ranking}
  \begin{tabular}{cr}
      \toprule
      Team Name & Accuracy score \\
      \midrule
      \textbf{Meta\_Learners} & \textbf{0.4042} \\
      ctom & 0.3573 \\
      Edinburgh & 0.2889 \\
      \bottomrule
  \end{tabular}
\end{table}

\begin{table*}[ht]
  \centering
  \caption{5-way 1-shot experiment results of different meta-learning methods on various few-shot datasets. The ResNet50 backbones load the pretrained weights from ImageNet.}
  \label{tab:main_results}
  \begin{tabular}{cc|rrrr}
      \toprule
      \textbf{Meta-learner} & \textbf{Backbone} & \textbf{Omniglot} & \textbf{CIFAR-100} & \textbf{miniImageNet} & \textbf{tieredImageNet}\\
      \midrule
      MAML & CifarCNN & \textbf{94.45} & 41.78 & 36.29 & 35.64 \\
      MetaCurvature & CifarCNN & 92.43 & 40.11 & 36.37 & 40.22 \\
      ProtoNet & CifarCNN & 63.26 & 35.47 & 30.48 & 31.05 \\
      ProtoNet & ResNet50 & 68.35 & 41.28 & 34.16 & 39.78 \\
      \midrule
      Base & ResNet50 & 91.54 & 49.59 & 41.32 & 48.41 \\
      Base+MCT & ResNet50 & 92.77 & 50.61 & 42.79 & 49.27 \\
      Base+Rot+MCT & ResNet50 & 89.95 & 52.72 & 43.31 & 49.72 \\
      MetaDelta (Ours) & Multiple & 93.56 & \textbf{56.83} & \textbf{47.52} & \textbf{51.74} \\
      \bottomrule
  \end{tabular}
\end{table*}

\begin{table}[ht]
  \centering
  \caption{Statistics of the five datasets for offline evaluation. The last three columns refer to the number of meta-train, meta-valid, and meta-test classes, and the number in brackets are the counts of super-classes. Partitions based on super-classes make the meta-learning more challenging.
  }
  \label{tab:dataset_statistics}
  \begin{tabular}{crrr}
      \toprule
      \textbf{Dataset} & \textbf{\# Meta-tr} & \textbf{\# Meta-val} & \textbf{\# Meta-te}\\
      \midrule
      Omniglot & 882 (25) & 81 (5) & 659 (20) \\
      CIFAR-100 & 50 & 10 & 40 \\
      miniImageNet & 50 & 10 & 40 \\
      tieredImageNet & 350 (10) & 56 (2) & 167 (8) \\
      \bottomrule
  \end{tabular}
  \vspace{-1mm}
\end{table}

\subsection{Offline Evaluation Results}

To evaluate the proposed MetaDelta in an offline environment, we conduct experiments on four public datasets for the few-shot image classification task and compare MetaDelta with several meta-learning baselines.

\subsubsection{Datasets}

Besides the public offline dataset Omniglot \cite{lake2015human} provided by the Challenge, we also select three popular few-shot image classification datasets to evaluate the generalization capacity of MetaDelta. The datasets include CIFAR-100 \cite{krizhevsky2009learning}, miniImageNet \cite{vinyals2016matching}, and tieredImageNet \cite{ren2018meta}. For each dataset (except for the officially provided Omniglot), we randomly partition the classes into meta-train, meta-valid, and meta-test sets according to the ratio of 5:1:4. 
The statistics of the five datasets is demonstrated in Table~\ref{tab:dataset_statistics}. Note that the images in all datasets are reshaped to $28\times 28$ to be consistent with the official interface (not included in the time budget, as the official images are claimed to always be of this size).

\subsubsection{Baselines}

Several representatives of optimization-based and metric-based meta-learning methods are adopted as our baselines. For optimization-based methods, we select MAML~\cite{finn2017model} and MetaCurvature~\cite{park2019meta}, an enhanced version of MAML that transforms the inner-update gradients to improve generalization capacity. CifarCNN is chosen as the backbone (base learner) in MAML and MetaCurvature due to its effectiveness and efficiency - larger backbones like (pretrained) ResNet50 cannot converge to good optimum within the time limit of the competition. For metric-based methods, we select ProtoNet~\cite{snell2017prototypical} with CifarCNN and pretrained ResNet50 backbones as our baselines, which applies episodic training rather than batch training as in MetaDelta. Moreover, the following variants of MetaDelta are adopted as baselines to show the impact of different components: 1) \emph{Base:} It fine-tunes a pretrained ResNet50 by batch training and makes inference with ProtoNet decoder during meta-testing. 2) \emph{Base+MCT:} It adopts MCT decoder for accurate prediction during meta-testing. 3) \emph{Base+Rot+MCT:} This baseline further applies the rotation loss augmentation in the batch training. 4) \emph{MetaDelta:} This is our final system with meta-ensemble on the predictions of four meta-learners. 

\subsubsection{Results}

The performance comparison is listed in Table~\ref{tab:main_results}. The proposed MetaDelta significantly outperforms other baselines on all datasets except Omniglot. 
However, we do not adopt MAML-based methods in our system as one of the meta-learners due to their low performance on the majority of datasets. As shown in Table~\ref{tab:main_results}, the meta-learner adopted in MetaDelta (\emph{Base} and the following variants) surpasses the typical MAML- and ProtoNet-based baselines by a large margin on CIFAR-100, miniImageNet, and tieredImageNet, and the MCT decoder with rotation loss augmentation help to boost the performance. 

{\bf Why ProtoNet Better} We notice in our experiment that, in almost all the datasets, our ProtoNet baseline (Base) and its variants outperform MAML by a large margin. We suspect that this superiority of ProtoNet-like methods is derived from the implicit utilization of the prior
knowledge of image data (e.g., locality, translation invariance, etc.) when combining pre-trained encoders with distance based decoders. In particular, most ProtoNet-like methods are specifically designed for few-shot image classification tasks,
projecting the images into latent vectors and making inferences based on the pairwise distances. On the other hand,
MAML-like methods adopt a more general framework without any assumption on the data or tasks, being not capable of leveraging these prior knowledge.

\subsubsection{Ablation Study}

We further conduct several ablation experiments to demonstrate the functionality of backbones and parameter-free decoders. Concretely, we implement single meta-learners with the backbones of ResNet50, ResNet152, WRN50 and MobileNet, and apply decoders of ProtoNet~\cite{snell2017prototypical} (\emph{Euclidean}), \emph{MCT}~\cite{kye2020transductive}, \emph{Laplacian}~\cite{ziko2020laplacian}, and Graph Propagation (\emph{Graph})~\cite{rodriguez2020embedding}. 

\begin{table}[ht]
  \centering
  \caption{Comparison of different backbones (pretrained CNN encoders). The Meta-learner is \emph{Base+Rot+MCT}.}
  \label{tab:backbone_cmp}
  \begin{tabular}{crr}
      \toprule
      \textbf{Backbone} & \textbf{CIFAR-100} & \textbf{tieredImageNet} \\
      \midrule
      ResNet50 & \textbf{52.72} & 49.72 \\
      ResNet152 & 51.06 & 49.25 \\
      MobileNet & 51.48 & 45.93 \\
      WRN50 & 49.69 & \textbf{50.14} \\
      \bottomrule
  \end{tabular}
  \vspace{-1mm}
\end{table}

Table~\ref{tab:backbone_cmp} lists the ablation results on the pretrained backbones, indicating that different backbones show superiority on different datasets. This observation motivates our design of taking different backbones in the four meta-learners and applying the meta-ensemble module, which aims at improving the generalization capacity of MetaDelta to unknown feedback datasets.

\begin{table}[ht]
  \centering
  \caption{Comparison of different decoders on the meta-test accuracies of CIFAR-100. The backbone is ResNet50.}
  \label{tab:decoder_cmp}
  \begin{tabular}{cr}
      \toprule
      \textbf{Decoder} & \textbf{CIFAR-100} \\
      \midrule
      MCT & \textbf{52.72} \\
      Euclidean & 50.71 \\
      Graph & 52.54 \\
      Laplacian & 51.37 \\
      \bottomrule
  \end{tabular}
\end{table}

\begin{figure}[ht]
  \centering
  \includegraphics[width=0.7\linewidth]{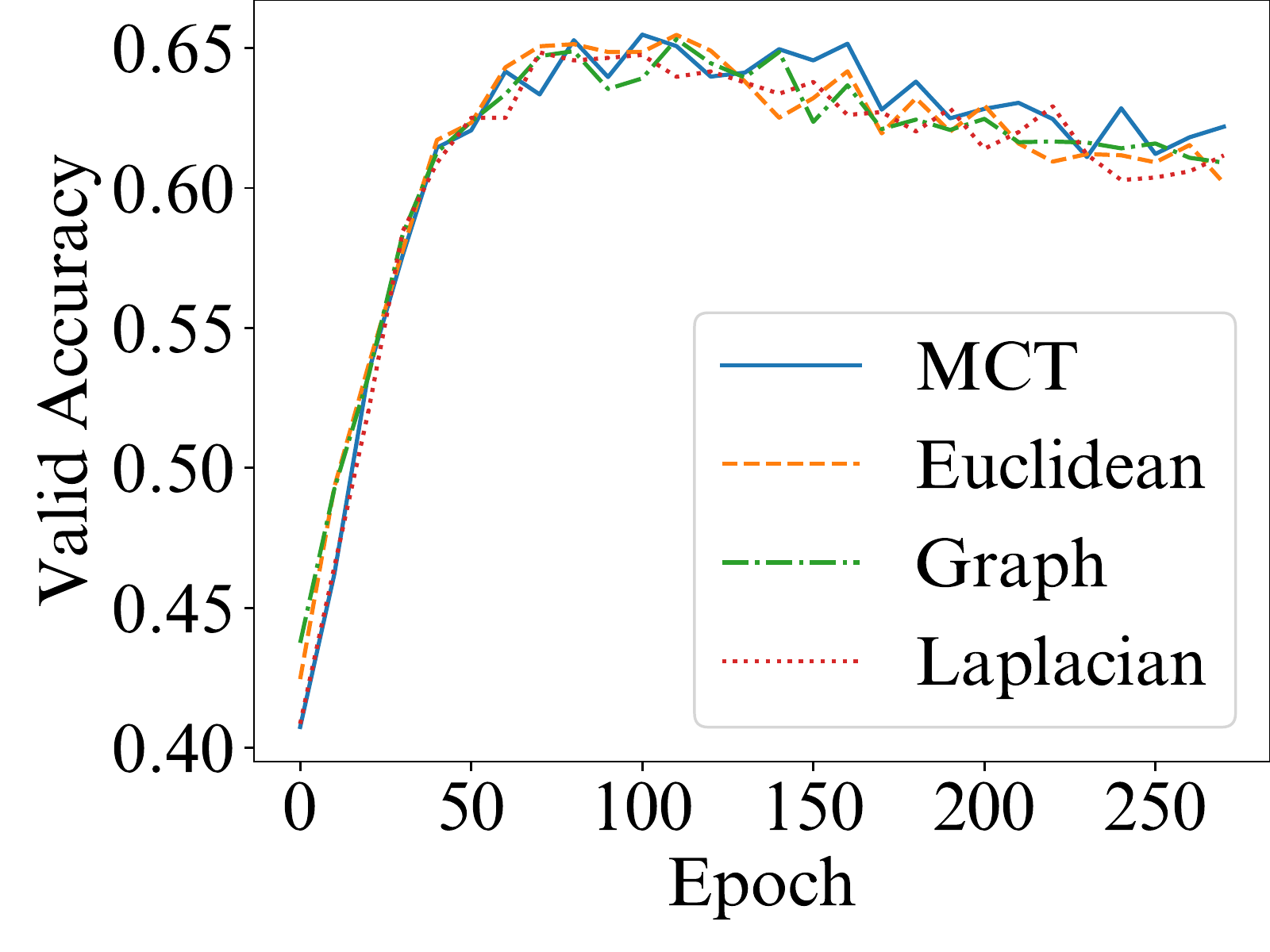}
  \caption{Comparison of different decoders on the meta-valid accuracies of CIFAR-100 dataset.}
  \label{fig:decoder_cmp}
\end{figure}

Table~\ref{tab:decoder_cmp} and Fig~\ref{fig:decoder_cmp} demonstrates the ablation results of different decoders in meta-learners. We can observe that the trending curves of the meta-valid accuracies of different decoders are akin to each other, which indicates that the best models saved according to Euclidean decoder and MCT decoder during meta-validation are the same with high probability. Therefore, we apply the Euclidean decoder during meta-validation for acceleration and the MCT decoder during meta-testing for higher accuracies (as shown in Table~\ref{tab:decoder_cmp}).

\section{Conclusion}

In this paper, we propose MetaDelta, a meta-learning system for few-shot image classification, which tackles two challenges of practical significance: 1) time and resource efficiency, and 2) generalization to unknown feedback datasets. For meta-learners in MetaDelta, we adopt a pretrained CNN encoder fine-tuned by batch training and a parameter-free decoder for inference. The meta-training of multiple meta-learners is arranged by a central controller with multi-processing techniques and a meta-ensemble module is applied to integrate the predictions. The resulting system ranks first in the final phase of the AAAI 2021 MetaDL Challenge. For future work, we plan to apply the domain generalization techniques~\cite{carlucci2019domain, li2019episodic} in computer vision to further enhance the generalization capacity of MetaDelta to any unknown datasets.

\section*{Acknowledgement}
This research is supported by the National Key Research and Development Program of China (No.2020AAA0106300, 2020AAA0107800,  2018AAA0102000) and National Natural Science Foundation of China No.62050110.

\bibliography{main}

\end{document}